\documentclass[iicol,sn-nature,pdflatex]{sn-jnl}

\usepackage{graphicx}
\usepackage{multirow}
\usepackage{amsmath,amssymb,amsfonts}
\usepackage{amsthm}
\usepackage{mathrsfs}
\usepackage[title]{appendix}
\usepackage{xcolor}
\usepackage{textcomp}
\usepackage{manyfoot}
\usepackage{booktabs}
\usepackage{algorithm}
\usepackage{algorithmicx}
\usepackage{algpseudocode}
\usepackage{listings}

\usepackage{xcolor} 
\newcommand\rev[1]{\textcolor{red}{#1}}
\usepackage[normalem]{ulem}
\renewcommand\rev[1]{#1}
\renewcommand\sout[1]{}

\DeclareMathOperator*{\argmin}{argmin}

\unnumbered

\begin{document}

\title[Beyond Knowledge Silos: Task Fingerprinting]{Beyond Knowledge Silos: Task Fingerprinting for Democratization of Medical Imaging AI}


\author*[1,2,3,4]{\fnm{Patrick} \sur{Godau}}\email{patrick.godau@dkfz-heidelberg.de}

\author[2,3]{\fnm{Akriti} \sur{Srivastava}}\nomail

\author[5,6]{\fnm{Constantin} \sur{Ulrich}}\nomail

\author[2,3]{\fnm{Tim} \sur{Adler}}\nomail

\author[1,3,4,5,6,7]{\fnm{Klaus} \sur{Maier-Hein}}\nomail

\author[1,2,3,6]{\fnm{Lena} \sur{Maier-Hein}}\nomail

\affil[1]{\orgname{National Center for Tumor Diseases (NCT)}, \orgdiv{NCT Heidelberg, a partnership between DKFZ and University Hospital Heidelberg},\orgaddress{\country{Germany}}}

\affil[2]{\orgname{German Cancer Research Center (DKFZ) Heidelberg}, \orgdiv{Division of Intelligent Medical Systems}, \orgaddress{\country{Germany}}}

\affil[3]{\orgdiv{Faculty of Mathematics and Computer Science}, \orgname{Heidelberg University}, \orgaddress{\country{Germany}}}

\affil[4]{\orgname{HIDSS4Health - Helmholtz Information and Data Science School for Health}, \orgaddress{\city{Karlsruhe/Heidelberg}, \country{Germany}}}

\affil[5]{\orgname{German Cancer Research Center (DKFZ) Heidelberg}, \orgdiv{Division of Medical Image Computing}, \orgaddress{\country{Germany}}}

\affil[6]{\orgdiv{Medical Faculty}, \orgname{Heidelberg University}, \orgaddress{\country{Germany}}}

\affil[7]{\orgdiv{Pattern Analysis and Learning Group}, \orgname{Heidelberg University Hospital}}

\abstract{The field of medical imaging AI is currently undergoing rapid transformations, with methodical research increasingly translated into clinical practice. Despite these successes, research suffers from knowledge silos, hindering collaboration and progress: Existing knowledge is scattered across publications and many details remain unpublished, while privacy regulations restrict data sharing. In the spirit of democratizing of AI, we propose a framework for secure knowledge transfer in the field of medical image analysis. The key to our approach is dataset ”fingerprints”, structured representations of feature distributions, that enable quantification of task similarity. We tested our approach across 71 distinct tasks and 12 medical imaging modalities by transferring neural architectures, pretraining, augmentation policies, and multi-task learning. According to comprehensive analyses, our method outperforms traditional methods for identifying relevant knowledge and facilitates collaborative model training. Our framework fosters the democratization of AI in medical imaging and could become a valuable tool for promoting faster scientific advancement.}




\maketitle

\section{Main}\label{sec:introduction}

\begin{figure*}[ht!]
\centering
\includegraphics[width=\textwidth]{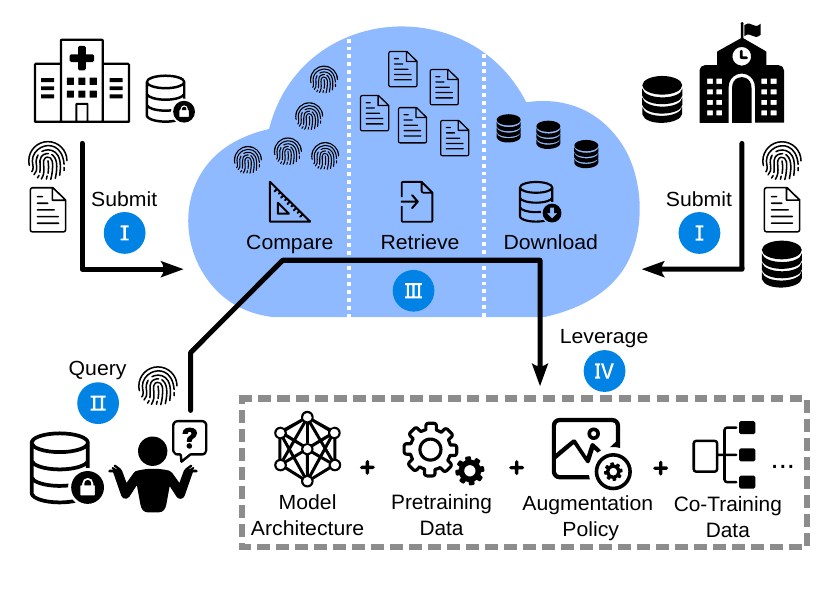}
\caption{\textbf{Concept of task fingerprinting for knowledge transfer.} (I) Participants of the knowledge cloud contribute by submitting a shareable task representation ("fingerprint"), meta information about their training strategies, and, optionally, their data. (II) A user generates the fingerprint for their task and queries the knowledge cloud. (III) Based on the most relevant tasks in the pool according to fingerprint matching, relevant training strategies and data can be retrieved. (IV) The retrieved meta information and data are used to compile a training pipeline with different components of transferred knowledge. In this study, we investigate four scenarios of knowledge transfer, namely (a) model architecture, (b) pretraining data, (c) augmentation policy, and (d) co-training data.
}\label{fig:concept}
\end{figure*}

Learning in isolation presents a substantial challenge in several domains of machine learning (ML). Particularly in the domain of medical imaging artificial intelligence (AI), strict regulations and high annotation costs incentivize the compilation of individual, private datasets, resulting in stakeholders creating independent \textbf{data silos}. In analogy, knowledge about successful deep learning training pipelines (including those on private data) often remains unpublished or dispersed across various publications and code repositories, making it difficult for individual researchers to effectively access and leverage past findings. The resulting partitioned landscape of \textbf{knowledge silos}, limited to individual researchers and labs, hinders overall scientific progress and puts researchers in the field at a mutual disadvantage. A recent study \cite{Eisenmann2023WhyIT} impressively showcased this issue, reporting that researchers spend a significant amount of time selecting model architectures and augmentation strategies for each new task, underscoring the potential of improved collaboration and automated knowledge transfer in advancing the field.

\begin{figure*}[ht!]
\centering
\includegraphics[width=\textwidth]{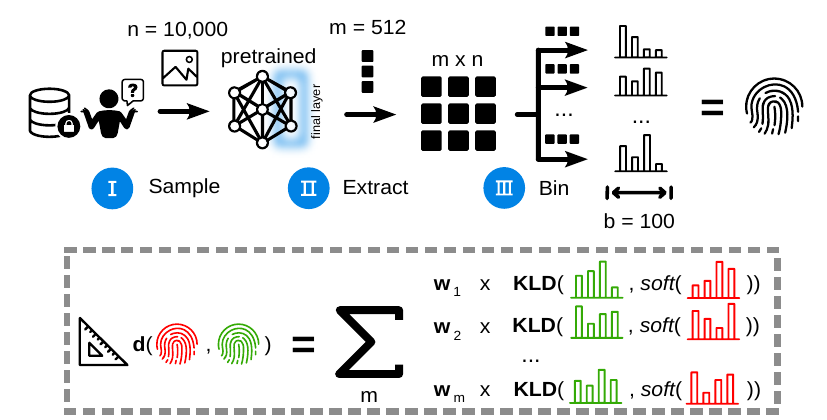}
\caption{\textbf{Proposed fingerprinting strategy and associated task distance measure: binned Kullback-Leibler Divergence (bKLD).} \rev{\sout{\textbf{Top}:}} To compute a task fingerprint, $n=10,000$ images are sampled (I) and passed through a pretrained backbone (we use an ImageNet \cite{Deng2009ImageNetAL} ResNet34 \cite{He2016DeepRL}) to extract $m=512$ features per image (II). The resulting $m \times n$ features are binned into $b=100$ bins along the features axis (III), resulting in m (normalized) histograms representing the fingerprint. \rev{\sout{\textbf{Bottom}:}} To compare two fingerprints, a weighted sum of the Kullback-Leibler Divergence (KLD) across all per-feature histograms is computed. A softmax operation is applied to source task histograms to avoid empty bins.}\label{fig:bkld}
\end{figure*}

The concept of \textbf{Transfer Learning} dates back to the 1970s \cite{Bozinovski2020ReminderOT} and involves any knowledge transfer from a source task to a target task \cite{Pan2010ASO}, with its most prominent application being pretraining on large-scale datasets \cite{Raghu2019transfusion,Zhuang2021ACS}. While utilizing off-the-shelf pretrained models is a common practice to speed up model convergence \cite{Raghu2019transfusion}, the vast number of available architectures, pretraining schemes, and datasets, makes selecting the most suitable option a time-consuming process. This premise also holds for the nascent field of foundation models \cite{Bommasani2021OnTO}. It has been shown that simply relying on benchmarks for pretrained models, such as, ImageNet \cite{Deng2009ImageNetAL}, does not translate well to the medical domain \cite{Raghu2019transfusion}. This fueled research of \textbf{Task Transferability Estimation (TTE)}, which aims to quantify the potential of knowledge transfer between tasks \cite{achille2019information}. While the research problem has been known for decades  \cite{BenDavid2006AnalysisOR} and substantial progress has been made, challenges with respect to data privacy considerations, robustness in realistic heterogeneous data settings, and the avoidance of negative transfer remain \cite{Zhang2020ASO,Zhuang2021ACS,Ding2024WhichMT}. Existing transferability estimation methods, which attempt to assess model suitability for a target task, often have only been applied in unrealistically homogeneous settings \rev{\cite{Nguyen2021similarity,Dias2021ImageDataset2VecAI,achille2019task2vec,liu2025wasserstein}, lack large-scale validation (12 tasks in \cite{MolinaMoreno2023AutomatedSS}, nine tasks in~\cite{hatakeyama2024transferability}, eight tasks in~\cite{tan2021otce}, seven tasks in \cite{Ramtoula2023VisualDR}, five tasks in~\cite{alvarez2020geometric}), or are incompatible with data privacy requirements \cite{You2021LogMEPA}. Others are not suited for scalability due to computational complexity \cite{Cheplygina2017ExploringTS,Fifty2021EfficientlyIT,Tran2019TransferabilityAH,tan2021otce,peng2020domain2vec,nguyen2020leep} or the assumption that tasks share underlying images \cite{Zamir2018TaskonomyDT}.}

\begin{figure*}[hp!]
\centering
\includegraphics[width=0.91\textwidth]{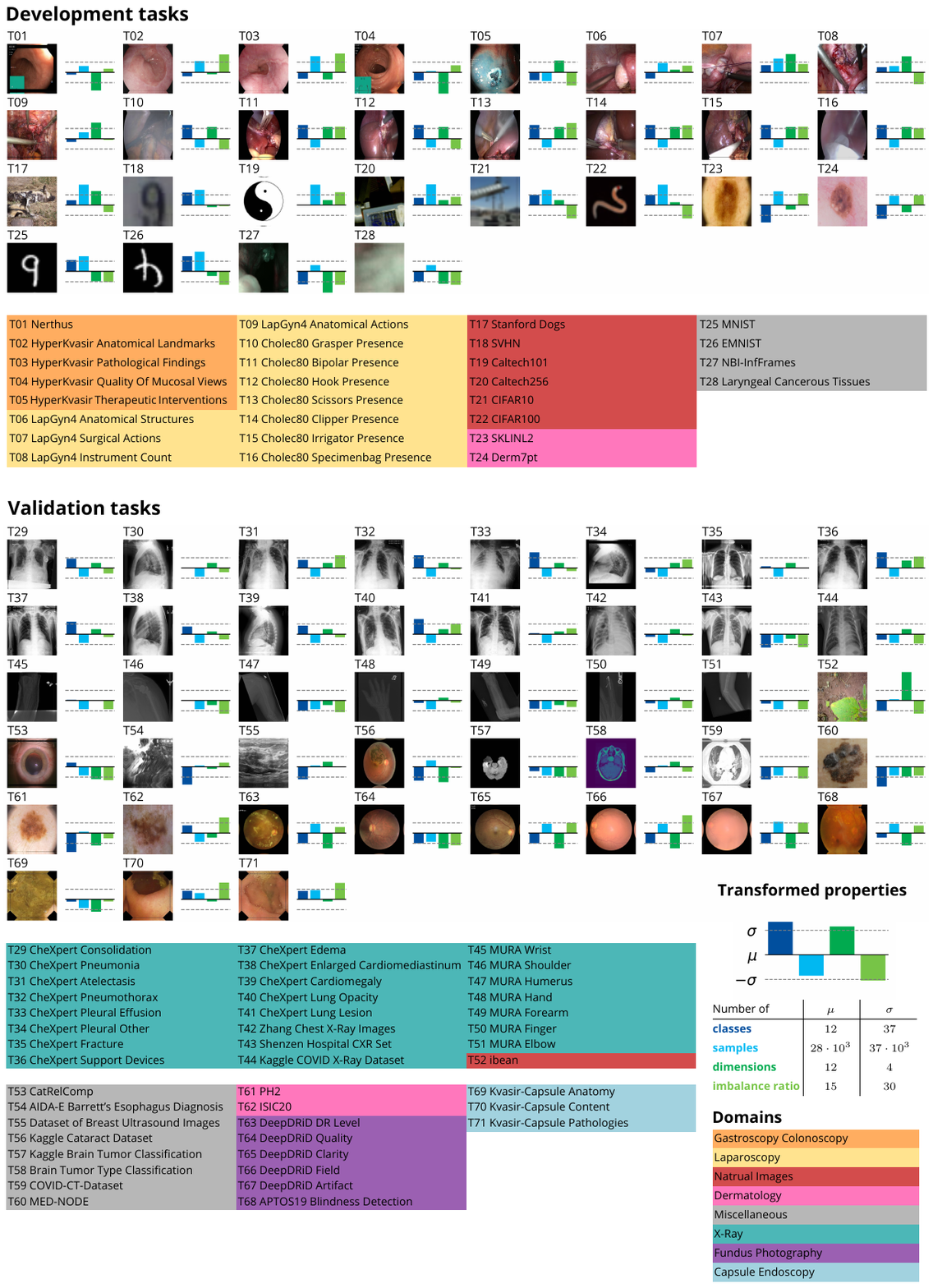}
\caption{\textbf{Our study is based on a heterogeneous set of 71 imaging tasks.} For each of the 28 tasks of the development split as well as 43 tasks of the validation split, we show a sample image next to the distribution of the following Box-Cox and Z-score transformed properties: number of classes, number of samples, intrinsic data dimension \cite{Pope2021dimension}, and imbalance ratio (size of largest class divided by size of the smallest class). The imaging domain is encoded as the background color of the dataset name.}\label{fig:data}
\end{figure*}

The core challenge of TTE lies in efficient task matching for knowledge transfer. Selecting an inappropriate source task for a new task can lead to detrimental negative transfer \cite{Wang2018CharacterizingAA,Zhang2020ASO}. Effective knowledge transfer requires generalizability across multiple components of the training pipeline - not just on pretraining, an aspect particularly underexplored in previous research. Additionally, creating effective task representations for knowledge transfer necessitates a balance between size (to facilitate sharing), data privacy (sufficient anonymization), and information content (retaining relevant task knowledge).

\begin{figure*}[h!t]
\centering
\includegraphics[width=\textwidth]{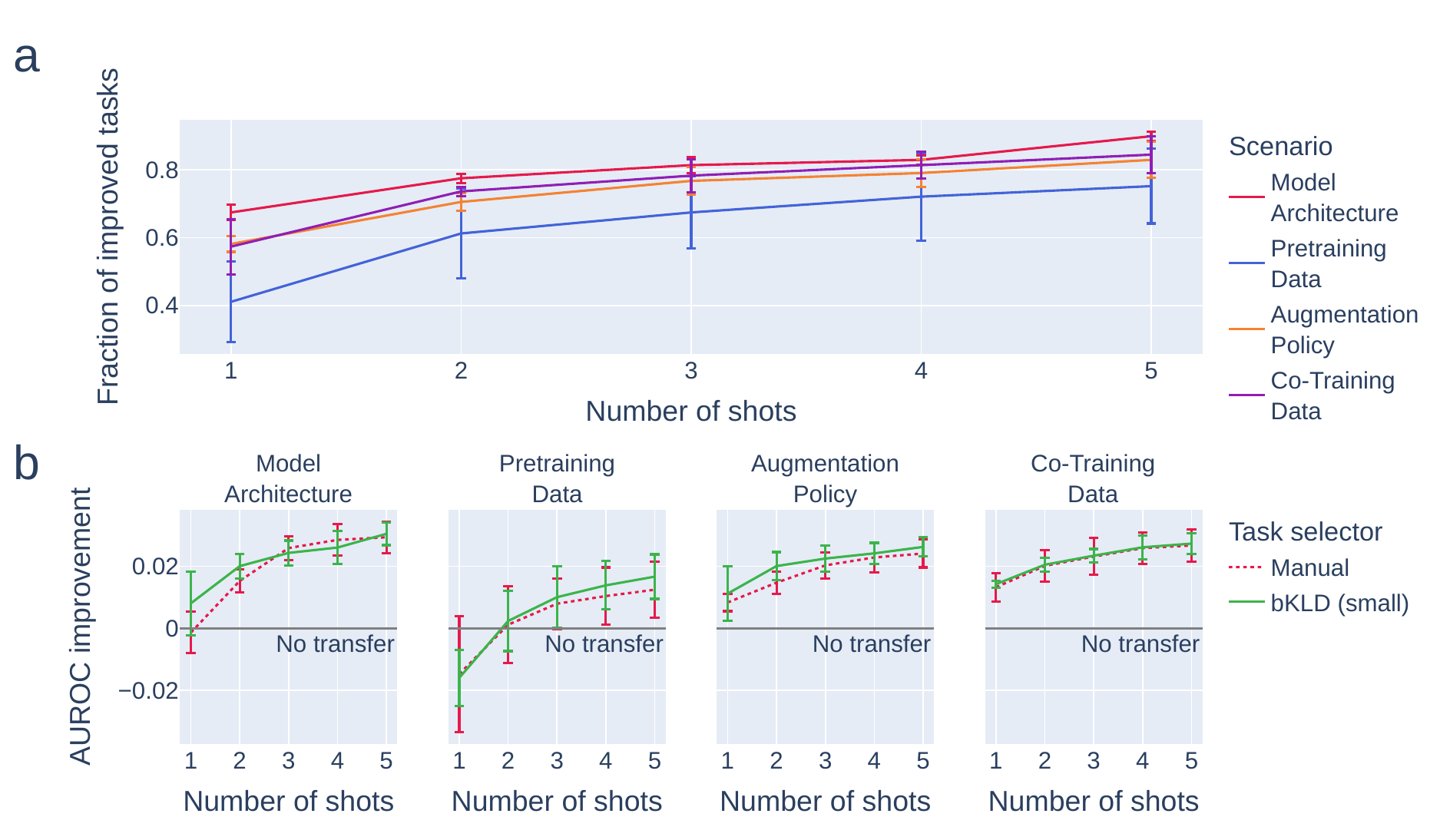}
\caption{\textbf{Task fingerprinting benefits training pipeline configuration and beats manual knowledge transfer.} \rev{\textbf{(a) \sout{Top}}}: Fraction of n=43 validation tasks that improve Balanced Accuracy (BA) through knowledge transfer ("gain" \cite{Zamir2018TaskonomyDT}) in four scenarios. \rev{\textbf{(b) \sout{Bottom}}}: Average delta in Area Under the Receiver Operator Characteristic (AUROC) across n=43 validation tasks. X-axis shows the number of shots, translating to the best of top k suggestions of our framework. Error bars correspond to standard deviation over three repetitions of all model trainings. Our proposed binned Kullback-Leibler Divergence (bKLD) fingerprint (here: the small variant) improves training for up to 90\% of validation tasks.}\label{fig:benefits}
\end{figure*}

To address these issues, we propose a novel, comprehensive framework for privacy-aware, collaborative knowledge transfer in medical imaging AI (see Fig. \ref{fig:concept}). At the core of our approach lies a knowledge cloud, storing experiences of model training along with an encapsulated representation of the underlying data, which we call the \textbf{task fingerprint}. Contributors may submit training meta information and voluntarily share the underlying data. A user may query the system for existing relevant knowledge or data by generating the fingerprint of their own task, leveraging the retrieved information during the compilation of their training pipeline. The key component of our approach is a new, efficient, and generalizable task distance measure computed from the individual task fingerprints.  We refer to this distance measure as \textbf{binned Kullback-Leibler Divergence (bKLD)} (see Fig. \ref{fig:bkld}), as fingerprints are generated by binning extracted image features, with similarity calculated using a weighted variant of the Kullback-Leibler Divergence. We demonstrate the effectiveness of our framework and task distance measure on the largest ever utilized set of heterogeneous tasks (n=71) in the context of medical imaging (see Fig. \ref{fig:data}). We combine this scale in task variety with diverse knowledge transfer scenarios (n=4), namely transferring the underlying model architecture, using source task data for pretraining, leveraging automatically generated augmentation policies, and simultaneously learning a co-task. The code, data, and intermediate results of our experiments are available at \url{https://github.com/IMSY-DKFZ/task-fingerprinting}. All models and datasets are publicly available.

\begin{figure*}[hp!]
\centering
\includegraphics[width=0.95\textwidth]{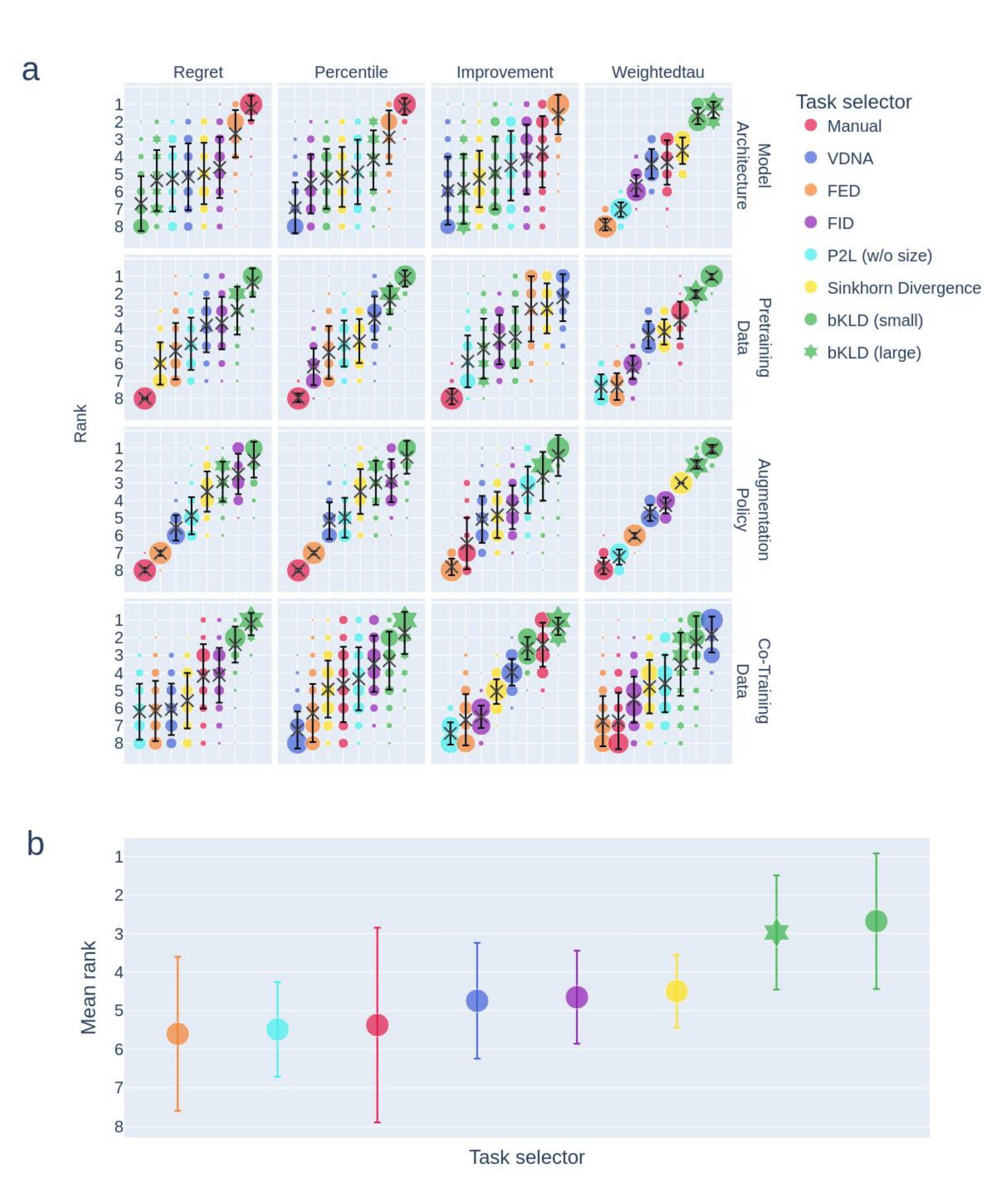}
\caption{\textbf{binned Kullback-Leibler Divergence (bKLD) outperforms previously proposed methods for knowledge transfer.} Uncertainty-aware ranking of our proposed bKLD methods for task fingerprinting versus \rev{manual task selection, VisualDNA (VDNA) \cite{Ramtoula2023VisualDR}, Fisher Embedding Distance (FED) \cite{Godau2021TaskFing,achille2019task2vec}, Fr\'echet Inception Distance (FID) \cite{Heusel2017FID,Ding2021AnalyzingDN}, Predict To Learn (P2L) \cite{Bhattacharjee2020P2LPT}, and Sinkhorn Divergence \cite{feydy2019interpolating}}. \rev{\textbf{(a)\sout{Bottom}}}: Columns represent four meta metrics to compare distance measures, whilst rows correspond to four knowledge transfer scenarios. We average across the top three suggestions by each method (weightedtau evaluates the full candidate ranking). Blob size shows frequency of rank across 1000 bootstrap samples from 258 setups (2 base metrics, 43 validation tasks, 3 repetitions). X marks the mean rank and whiskers the standard deviation. Plot is inspired by \cite{Wiesenfarth2021RR} and follows the “aggregate then rank" assessment method. \rev{\textbf{(b)\sout{Top}}}: Summary of the 16 subplots below. The marker position refers to the mean rank over individual bootstrapped mean rankings (X marks) and whiskers indicate standard deviation.}\label{fig:comparison}
\end{figure*}

\section{Results}\label{sec:results}

For comprehensive validation, we generated a knowledge cloud comprising a pool of 71 tasks as shown in Fig. \ref{fig:data}. To avoid overfitting our task fingerprints, only 28 tasks were used for development and hyperparameter tuning. The remaining 43 downstream tasks were used for prospective testing and covered the medical domains of radiology, dermatology, ophthalmology, endoscopy, and sonography.

To investigate the impact of knowledge transfer on complementary modules of the training pipeline, we queried the knowledge cloud for relevant existing metadata, such as the best-performing model architecture and suitable augmentation policy, or for additional data samples for pretraining and co-training. As we repeated every possible knowledge transfer three times to compensate for non-determinism during model training, we trained more than 30,000 neural networks solely for the evaluation part, which resulted in roughly 10,000 GPU hours of training.

\subsection{Task fingerprinting enables knowledge transfer without sharing sensible data}\label{ssec:res:benefits}

For a fraction of 67\% of validation tasks, the resulting models exhibited an improvement by switching our baseline model architecture to the prime candidate proposed by the knowledge cloud. Replacing the augmentation strategy benefited 58\% of tasks' first shot, i.e., only leveraging the augmentation strategy derived from the best matching source task in the knowledge cloud. Using shared image data of the best-aligned source task for knowledge transfer through pretraining and co-training resulted in an improvement for 41\% and 57\% of validation tasks. Increasing the computational budget during downstream model training and considering multiple knowledge source candidates ("multi-shot") led to up to 90\% of validation tasks being improved, as shown in Fig. \ref{fig:benefits} (\rev{a \sout{top}}). Notably, this held true when transferring model architecture (90\%) and augmentation policy (83\%) solely via sharing the fingerprint as a compressed identifier without any other data samples leaving the institution of the data owner. Assuming that the task data was available for pre- or co-training, we observed a similar rate of task improvement for the co-training scenario (84\%), compared to our single task baseline. Considerably fewer tasks benefited from knowledge transfer via pretraining (75\%), while the fluctuations in our repeated experiments due to randomness were more severe.

\subsection{Task fingerprinting outperforms manual task selection}\label{ssec:res:manual}

Reviewing the literature to identify previously successful training recipes has proven to be the predominant strategy chosen by participants of biomedical challenges \cite{Eisenmann2023WhyIT}. This strategy, which we refer to as "manual" task matching, solely relies on the similarity of the semantic task description, is limited to the actual training information published on a similar task, and requires manual extraction from the publication or repository. Simulating such semantic task matching allowed us to compare our fully data-driven task fingerprinting with this prevailing strategy. The results are shown in Fig. \ref{fig:benefits} (\rev{b\sout{bottom}}), which displays the average performance increase per task compared to our single task baseline. The relative performance increase by task fingerprinting was 12\%, 57\%, 15\%, and 2\% larger than manual task matching for the investigated knowledge transfer scenarios of model architecture, pretraining, augmentation policy, and co-training, respectively, when averaging across all tasks, repetitions, and the multi-shot range from 1 to 5.

\subsection{Our proposed task distance measure outperforms previously suggested ones}\label{ssec:res:comparison}

To compare our fingerprinting technique to previously proposed task similarity measures, we instantiated our framework with previously proposed task transferability measures including VisualDNA (VDNA) \cite{Ramtoula2023VisualDR}, Predict To Learn (P2L) \cite{Bhattacharjee2020P2LPT}, Fisher Embedding Distance (FED) \cite{Godau2021TaskFing,achille2019task2vec}\rev{, Sinkhorn Divergence \cite{feydy2019interpolating},} and Fr\'echet Inception Distance (FID) \cite{Heusel2017FID,Ding2021AnalyzingDN}. Note that the underlying task representations differ for each of these, with FID and VDNA being comparable in size to our bKLD-based task fingerprints, while P2L is significantly smaller and FED \rev{as well as Sinkhorn being} larger. We compare them to \rev{two} variants (hyperparameter configurations) of our proposed distance measure bKLD, which obtain the top \rev{two} positions in the overall ranking (see Fig. \ref{fig:comparison} \rev{b \sout{top}}). \rev{\sout{The bottom of}} Fig. \ref{fig:comparison} \rev{(a)} shows a detailed analysis of the rankings for all mentioned methods with respect to meta metrics and knowledge transfer scenarios. \rev{\sout{Within all four knowledge transfer scenarios, for at least three out of four meta metrics, a single variant of bKLD performed best.}} Additional evaluation aggregation schemes can be found in the supplementary information (Supplementary Table \ref{tab:win_rates} and Supplementary Fig. \ref{fig:app_comparison}).

\begin{figure*}[ht!]
\centering
\includegraphics[width=\textwidth]{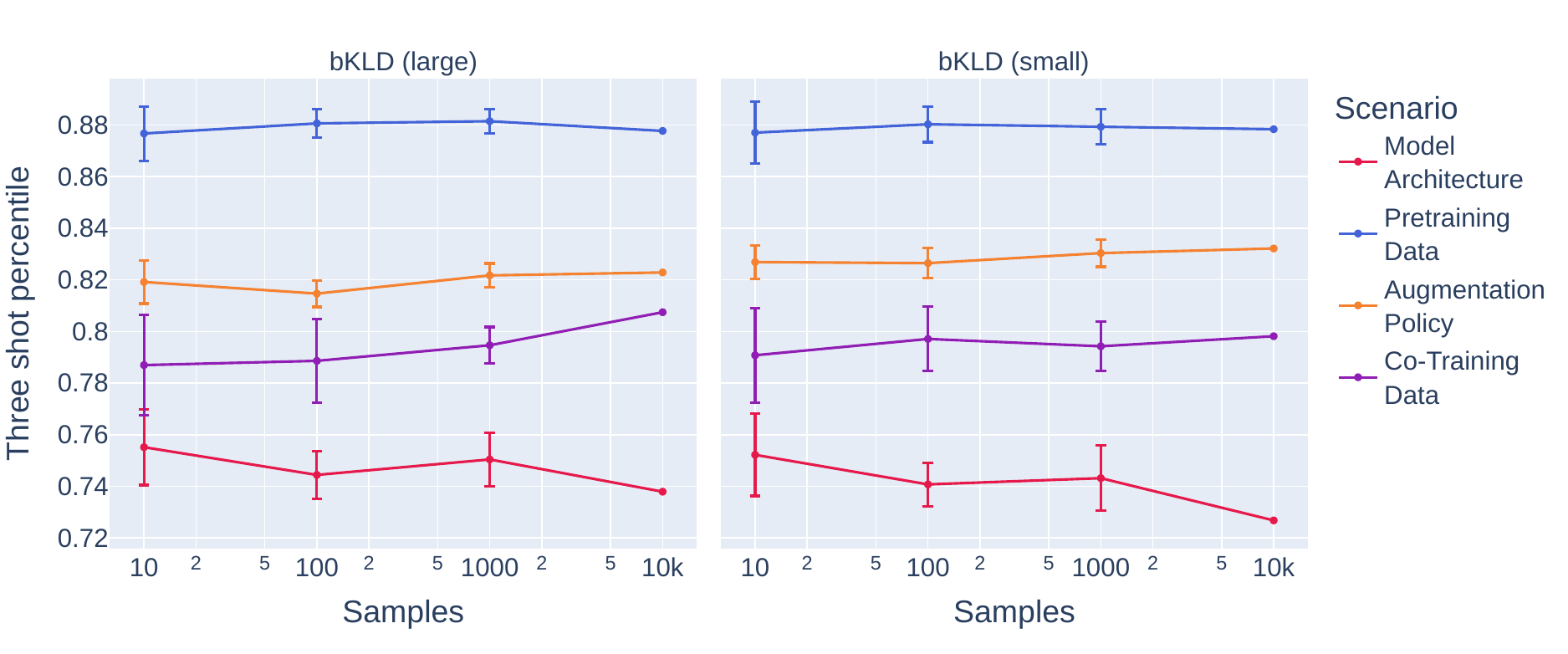}
\caption{\textbf{binned Kullback-Leibler Divergence (bKLD) is robust with respect to dataset size.} Percentile of best 3-shot source task knowledge transfer according to Balanced Accuracy (BA) averaged over n=43 validation tasks in four scenarios applying \rev{two} proposed bKLD variants. X-axis shows the number of samples used from a task to generate its fingerprint in log scale. Error bars indicate standard deviation over 10 resamplings.}\label{fig:robustness}
\end{figure*}

Comparing transferability measures has been shown to be a very sensitive experiment, where small variations in the setup may lead to conflicting conclusions on superiority \cite{Agostinelli2022HowStable}. Therefore, the \textbf{setup stability score} \cite{Agostinelli2022HowStable}, has been proposed to indicate how stable the ranking of task similarity algorithms is with respect to varying a single component of the setup, e.g. choosing a different performance metric. A value of 0 indicates no correlation between the rankings and 1 denotes perfectly identical rankings. For our experiments, the stability scores were \rev{0.59 \sout{0.61}} for the meta metrics, \rev{0.31 \sout{0.32}} for the performance metrics, \rev{0.1 \sout{0.12}} for the random seed, 0.02 for the target tasks, and only 0.01 for the knowledge transfer scenario.

\subsection{Our proposed task distance measure is robust with respect to task size}\label{ssec:res:robustness}

As fingerprint generation is a non-deterministic process and we envision our methodology to be especially useful for small tasks, we varied the number of samples used to compute a fingerprint. While all previous experiments (including all alternative task similarity measures) evaluated 10,000 samples from a task, we tested significantly reduced sizes of up to only 10 samples. Figure \ref{fig:robustness} shows our method’s stability in retrieving sufficiently beneficial task candidates in a best of three shots scenario, where random task selections would be expected to achieve a value of 0.75. Based on fingerprints generated from only 10 samples \rev{both bKLD variants} show robust task matching performance. Of note, knowledge transfer for pretraining data has the best \textit{relative} task matching of the four scenarios investigated, meaning that out of the available sources typically ones around the 88th percentile are chosen (Fig. \ref{fig:robustness}). This stands in contrast to the \textit{absolute} improvement to the validation tasks as shown in Fig. \ref{fig:benefits}, where applying pretraining knowledge transfer benefited the least across the scenarios investigated.

\begin{figure*}[ht!]
\centering
\includegraphics[width=\textwidth]{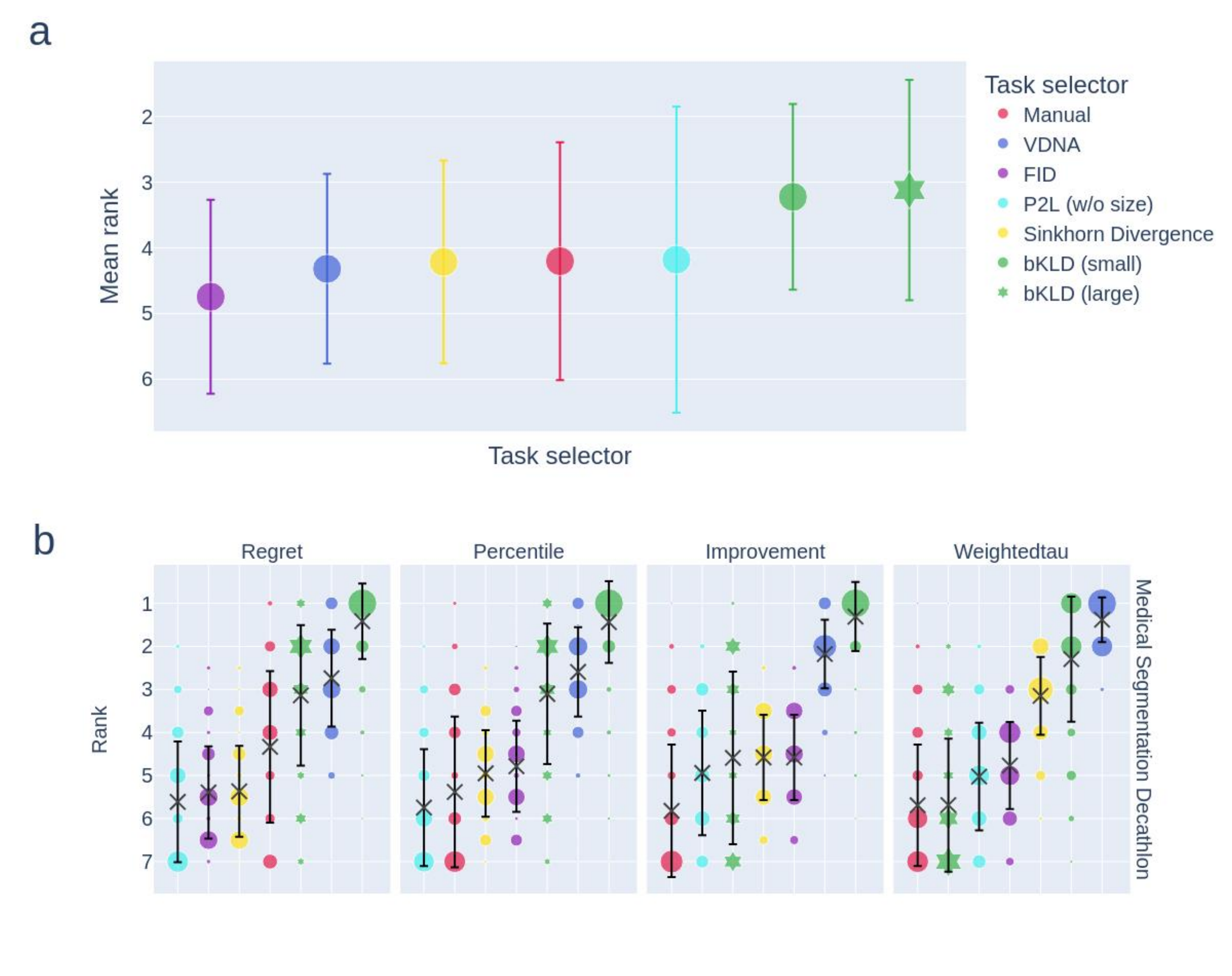}
\caption{\rev{\textbf{binned Kullback-Leibler Divergence (bKLD) generalizes to 3D, new task types and further backbone models.} Uncertainty-aware ranking for 2D classification and 3D medical image segmentation tasks demonstrating strong performance of our proposed bKLD methods for task fingerprinting compared to manual task selection, VisualDNA (VDNA) \cite{Ramtoula2023VisualDR}, Fr\'echet Inception Distance (FID) \cite{Heusel2017FID,Ding2021AnalyzingDN}, Predict To Learn (P2L) \cite{Bhattacharjee2020P2LPT}, and Sinkhorn Divergence \cite{feydy2019interpolating}. \textbf{(a)}: Extension of Fig.~\ref{fig:comparison} to three additional feature extraction backbones. Summary of 48 subplots (16 from a CLIP~\cite{radford2021learning}, DINO~\cite{caron2021emerging}, and MAE~\cite{he2022masked} backbone respectively). The marker position refers to the mean rank over individual bootstrapped mean rankings (X marks) and whiskers indicate standard deviation. \textbf{(b)}: Extension of Fig.~\ref{fig:comparison} to 3D medical imaging using data from the Medical Segmentation Decathlon~\cite{antonelli2022medical}. Columns represent four meta metrics to compare distance measures. We average across the top three suggestions by each method (weightedtau evaluates the full candidate ranking). Blob size shows frequency of rank across 1000 bootstrap samples from 34 setups (2 base metrics, 17 validation tasks). X marks the mean rank and whiskers the standard deviation. Plot is inspired by~\cite{Wiesenfarth2021RR} and follows the “aggregate then rank” assessment method. }
}\label{fig:generalization}
\end{figure*}

\subsection{\rev{Our proposed task distance measure generalizes to unseen task type and feature extractors}}\label{ssec:res:generalization}

\rev{
To further test the generalization capabilities of bKLD, we investigated two variations of our experiments. First, we applied task transferability estimation on the outcomes of the Medical Segmentation Decathlon~\cite{antonelli2022medical}. In this international competition 19 participating teams submitted segmentation models for 17 segmentation tasks on 10 datasets. Data were obtained from the two most common 3D medical imaging modalities, namely 3D magnetic resonance imaging (MRI) and computed tomography (CT). Segmentation tasks were highly diverse, ranging from the segmentation of large objects (e.g., liver) to the segmentation of tubular structures (e.g., vessels). Note that  the task type (segmentation versus classification), the modalities (2D versus tomographic 3D) and the target structures (e.g., vessels) all represent a severe shift compared to our previous experiments~\cite{simpson2019large}. The outcomes are shown in Fig.~\ref{fig:generalization} (b) and are consistent with our previous findings. Concretely, the small variant of bKLD performs best in three out of four meta metrics. Second, we replaced the feature extractor used in our previous experiments with three more recent models based on CLIP~\cite{radford2021learning}, DINO~\cite{caron2021emerging}, and MAE~\cite{he2022masked}. These experiments represent shifts in the distribution of the extracted features and consequently the resulting task fingerprints. Fig.~\ref{fig:generalization} (a) shows the aggregated pendant to Fig.~\ref{fig:comparison} (b), i.e., the ranking of fingerprinting methods across meta metrics, transfer scenarios, and feature extractors, and is consistent with our main experiments.}

\section{Discussion}\label{sec:discussion}

This work introduced task fingerprinting as a novel approach to AI democratization. These shareable task embeddings enable knowledge transfer estimation, allowing researchers to collaboratively collect and leverage insights during neural network experiments. In particular, our method introduces a new task similarity measure named \textbf{binned Kullback-Leibler Divergence (bKLD)}, which is both robust and efficient in computation, as well as flexible to adapt to various knowledge transfer scenarios. Importantly, it outperformed previous approaches in our experiments which were, to the best of our knowledge, conducted using the largest heterogeneous set of tasks in combination with the broadest evaluation across knowledge transfer scenarios in the domain of task transferability estimation research. The key insights drawn from this comprehensive analysis are:

\begin{enumerate}
    \item \textbf{Task fingerprinting allows for secure knowledge transfer.} Our findings demonstrate the potential of task fingerprinting as a method for knowledge sharing, particularly in the data-sensitive domain of medical imaging. Our approach improved downstream task performance for a majority of tasks within a reasonable computational budget, while outperforming the prevailing and time-consuming practice of manual knowledge compilation. 
    \item \textbf{The best fingerprinting strategy varies substantially across scenarios.} The reason we did not identify a one-size-fits-all solution  \rev{(see Fig. \ref{fig:comparison} a)} is that the optimal source task rankings are inconsistent  \rev{(see Supplementary Table~\ref{tab:scenario_similarity})}, hence  no single task distance measure might suit all transfer scenarios. \rev{Despite this, bKLD demonstrates strong generalization capabilities}. \rev{\sout{The weighting mechanism of bKLD solves this dilemma and shifts attention to features that are relevant for a given scenario.}} 
    \item \textbf{Task distance measures benefit from feature binning.} Our newly introduced task distance measure bKLD outperformed existing measures \rev{\sout{in all evaluated transfer scenarios}}. While the Predict To Learn (P2L) \cite{Bhattacharjee2020P2LPT} approach computes the Kullback-Leibler Divergence (KLD) on averaged features, the binning strategy in bKLD allows a much more granular comparison of image feature distributions. The introduced weighting mechanism further provides an option for prioritization of dominant features of the \rev{\sout{source or}} target task.
\end{enumerate}

\rev{\sout{We consider the broader implications of our research to be significant: The widespread adoption of knowledge clouds based on task fingerprinting has the potential to democratize AI research by facilitating collaboration and knowledge sharing. Shared knowledge, in turn, can lead to reduced model development times and a decrease in carbon emissions associated with extensive training processes.}}

While our method is straightforward to use, it comes with \rev{\sout{one} some} important hyperparameter\rev{s. The number of bins allows to control the tradeoff between representation granularity on the one hand and privacy and efficiency on the other hand. The} weighting mechanism \rev{allows to make use of the same representation in multiple ways, thus} enables bKLD to adapt to various scenarios of knowledge transfer. \rev{Importantly, the comparability of fingerprints requires an identical feature extraction model and at least a similar preprocessing method. Otherwise the $j$-th feature dimension of model $A$ is unlikely to be directly related to the $j$-th feature dimension of model $B$, and the total number of feature dimensions may differ. Although our experiments (Fig. \ref{fig:generalization}) demonstrate that bKLD generalizes well beyond a single feature extractor, major considerations regarding the maintainability of knowledge in the knowledge cloud remain. In theory, it would be feasible to support multiple fingerprinting pipelines (comprising preprocessing and feature extraction) in parallel for  images with widely varying natures (e.g., histopathology and 3D imaging) and occasionally update these pipelines to support multiple extraction models over time. This approach would provide sufficient flexibility to introduce more powerful feature extractors while avoiding complete  erasure of existing knowledge with the introduction of a single new extraction pipeline.}

\rev{\sout{Setting the weights of bKLD according to dominant source task features was especially beneficial in the model architecture scenario (see Fig. \ref{fig:comparison} bottom), though this variant of bKLD was the least robust with respect to image sampling (see Fig. \ref{fig:robustness}) and the scenario} Compared with previous methods, bKLD performed poor in the model architecture scenario (see Fig. \ref{fig:comparison} a), which also} obtained the lowest percentile score (see Fig. \ref{fig:robustness}). Furthermore, manual task selection suited this scenario better than all others, presumably indicating that granular feature distributions have a subordinate role in the fitting of model architectures. For this specific scenario, also known as \textbf{source-free model transferability estimation}, many solutions have been proposed, but the problem remains unsolved \cite{Ding2024WhichMT}. \rev{On a wider scope our generalization experiment on the Medical Segmentation Decathlon shows that the transfer of a complete model pipeline may be better predicted (Fig. \ref{fig:generalization} b).} For the pretraining scenario, \textbf{bKLD(small\rev{\sout{,target}})} proposed source tasks ranked highest over all scenarios (see Fig. \ref{fig:robustness}). Interestingly, in absolute terms, the performance improvement was low compared to the other training scenarios (see Fig. \ref{fig:benefits}), i.e., for pretraining the bKLD method appeared to perform best, but the improvement is still small compared to other scenarios. We interpret this as a lack of tasks that are suited well for pretraining in our experimental data pool, as pretraining tasks ideally are large in size, and comprise sufficient variety in sample distribution to cover the downstream task \cite{Kolesnikov2019BigT,Mensink2021FactorsOI}. \rev{\sout{The consistently poor performance of \textbf{bKLD(large,source)} in this scenario opposed to the other weighting strategies (see Fig. \ref{fig:comparison} bottom) further reveals the importance of feature distribution similarity especially with respect to dominant features of the target task.} Within the augmentation policy scenario \textbf{bKLD}(small) outperformed all other task selectors consistently on all meta metrics (see Fig. \ref{fig:comparison}), demonstrating the strength of its image representation.}
In the co-training scenario, where the unweighted \rev{large} bKLD variant performed best, it appears intuitive that focusing primarily on one task’s dominant features does not yield better transfer estimation. Though the \textbf{bKLD(small\rev{\sout{,target}})} and \textbf{bKLD(large\rev{\sout{,unweighted}})} variants’ meta metrics were very close to each other in most cases (see Fig. \ref{fig:comparison}, \ref{fig:robustness}, Supplementary Fig. \ref{fig:app_comparison}).

\rev{While the knowledge cloud may hide the actual task fingerprints from users by returning only distances and metadata, the central service will always see the full task fingerprints. To preserve privacy, it is necessary to sufficiently obfuscate patient information within the fingerprints to prevent attacks such as membership inference or data reconstruction \cite{el2024preserving}. \textbf{bKLD} offers three mechanisms to accomplish this: aggregation, decoupling, and quantization. The binning procedure is key here. In this step feature vectors of individual images are broken up (i.e., decoupled or shuffled), and each entry is assigned to separate bins (i.e., quantized or perturbed). Moreover, the final fingerprint consists of the aggregated histograms for all images. Although we do not provide full theoretical evidence, we can point to the literature, where these techniques are common tools in differential privacy \cite{kaissis2020secure}.}

\rev{Consideration of computational costs is highly relevant in the context of both democratization of AI and environmental impact. Related to this work, two different perspectives need to be distinguished: The resources for developing and validating the present approach and the resources required for using our method. The “brute force” approach of combining every potential knowledge transfer between two tasks of our task pool was required to assess the meta metrics weightedtau, percentile and (most of the time) regret. But these are not the costs incurred during regular use: Computing task fingerprints (i.e., feature extraction and binning) took us only about 2 minutes for all tasks together on a standard workstation with 12 CPU cores and a single RTX 3090. A regular participant has to compute only a single task fingerprint once, which stored as a numpy array, makes about 0.4 MB (100 bins) to 4.1 MB (1000 bins). For every conducted experiment during model development, the outcome and hyperparameters for the corresponding fingerprint can be submitted in an automated fashion, without any noticeable overhead. The most compute intensive part remains the task distance computation in the cloud during a query, where all previously submitted fingerprints need to be considered. Fortunately, computing B (number of bins) times the Kullback-Leibler Divergence is a “cheap” operation and the computations for each existing fingerprint can be parallelized. In our implementation, computing distances for all combinations of task pairs within the task pool of size 71 took about 4 seconds for the 1000 bins bKLD variant (including data transfer to GPU). For an actual query not all internal combinations must be computed, hence our measurement roughly corresponds to a usual query with a knowledge cloud of a couple thousand distinct task submissions (except for the GPU data transfer). This computation time scales linearly with the size of the knowledge cloud. So on standard consumer hardware and without further potentially existing optimizations, our implementation should already scale to 100,000s of distinct fingerprints.}

Closely related to task fingerprinting is the field of \textbf{Automated Machine Learning (AutoML)}, which aims to fully automate the configuration of ML algorithms, specifically steps such as data preprocessing, feature engineering, and hyperparameter optimization \cite{He2021automl}. It shares the goal of AI democratization by resolving tricky steps in ML and empowers non-experts to generate AI systems. However, AutoML often raises computational expenses, especially in approaches such as Neural Architecture Search (NAS) \cite{Dias2021ImageDataset2VecAI,Le2022ftdnas}, and, in contrast to our approach, lacks a collaborative mechanism. Similarly, frameworks such as \textbf{nnU-net} \cite{Isensee2020nnUNetAS} have shown promise in offering self-configuring solutions for biomedical image segmentation tasks. Although training adjustment in nnU-net is way faster and substantially less resource-intensive than traditional AutoML, it is limited to the static rules designed by the authors (i.e., no new knowledge is gained automatically over time), focuses on a specific task type (only segmentation) and excludes crucial aspects of the training pipeline (e.g., architecture template and data augmentation are fixed). \textbf{Federated learning} \cite{Li2021ASO}, on the other hand, offers a path for collaborative learning while preserving data privacy, but imposes additional burdens related to synchronization, trust establishment, and preparation \cite{Rieke2020federated}. Task fingerprinting combines the efficiency of nnU-net with the flexibility of AutoML, while leveraging collaborative efforts such as Federated Learning. Similarly to the latter, a general limitation of task fingerprinting is that\rev{\sout{, in a public setting,}} the system is vulnerable to malicious actors submitting false data \rev{in a public setting. This \sout{, requiring} requires} the development of preventive measures\rev{, such as detection algorithms \cite{zhang2022fldetector} or trust scores \cite{cao2021fltrust}}. In addition, the success of the overall concept relies on the willingness of researchers to share knowledge within the community. Perhaps most importantly and in contrast to all three mentioned approaches, however, our method has the potential to facilitate \textbf{Lifelong Learning} \cite{parisi2019lifelong} by continuously and automatically growing knowledge, as more data and tasks are incorporated over time. This capability could lead to long-term scalability of AI which would reduce energy consumption and carbon footprint of systems \cite{Soltoggio2024ACA}.

It is important to consider our findings in the context of previous research on task distance measures. In line with our preliminary work \cite{Godau2021TaskFing} the FED task selector ranks high for the model architecture and pretraining scenarios and some meta metrics. However, it falls short on the other meta metrics and knowledge transfer scenarios. Prior studies have also highlighted the inconsistency of transferability estimations in general \cite{Agostinelli2022HowStable} and specifically in the medical domain \cite{chaves2023performance}. This aligns well with our observations regarding the low stability scores of existing distance measures across target tasks and the transfer scenario and emphasizes the importance of finding a flexible approach that can adapt to multiple transfer scenarios. In fact, Standley et al. \cite{Standley2020WhichTS}, also observed that task similarity measures might be suited for one kind of knowledge transfer, e.g., pretraining, but inappropriate for another, e.g., co-training. \rev{From this perspective, the consistency with bKLD is remarkable. bKLD also excels as a task selector for 3D segmentation tasks outside of our original experimental design.} Our aggregation over multiple metrics and meta metrics was intended to counteract the sensitivity of evaluations to specific metrics \cite{reinke2024understanding} and increase the assessment robustness as well as allow for the consideration of multiple knowledge transfer scenarios. We observed that weightedtau allows for the most stable separation of task selectors, as opposed to regret, percentile, and improvement (Fig. \ref{fig:comparison} \rev{a \sout{bottom}},  Supplementary Fig. \ref{fig:app_comparison}). However, it is hard to interpret and circumvents a resource limitation-based cutoff.

The emergence of foundation models \cite{Bommasani2021OnTO} has amplified the importance of AI democratization, as the \rev{necessary \sout{requisite}} expertise, computational resources, and data volumes have grown substantially. \rev{These \sout{Such foundation}} models \rev{could \sout{might}} bridge the gap between different modalities in medical imaging through their large-scale pretraining \cite{Azad2023FoundationalMI}. However, \rev{\sout{they are accompanied by}} two fundamental problems \rev{accompany them}: first, \rev{selecting the most suitable one from \sout{navigating}} the diverse landscape of available \rev{ones \sout{models to select the most suitable one}}, and second, devising an effective adaptation strategy for specific downstream tasks. \rev{Although integrating \sout{While the integration of}} multiple modalities into fingerprints is \rev{a topic \sout{left}} for future research, we note that \rev{\sout{these challenges are well-suited to be answered by}} task fingerprinting \rev{is a promising approach to addressing these challenges}. \rev{Notably, the computational overhead of task fingerprinting – both in terms of archiving post-experimental knowledge and querying prospective knowledge – is minimal, especially compared to the actual model training process. Therefore, task fingerprinting has the potential to accelerate the selection and adaptation of foundation models.
}

The actual implementation of a knowledge cloud remains a future challenge for the realization of our concept. Using existing infrastructure, such as the “Joint Imaging Platform” \cite{scherer2020joint}, where cross-institutional research in the field of medical imaging is already being carried out, would provide an excellent starting point.
Additionally, as our experiments treated each transfer scenario in isolation, investigating the impact of transferring multiple components of a pipeline at the same time (entangled transfer scenarios) is a crucial next step. Expanding upon our evidence by testing other task types, a larger number of target tasks, and other parts of the training pipeline is also essential.

\rev{We consider the broader implications of our research to be significant: The widespread adoption of knowledge clouds based on task fingerprinting has the potential to democratize AI research by facilitating collaboration and knowledge sharing. Shared knowledge, in turn, can lead to reduced model development times and a decrease in carbon emissions associated with extensive training processes.}

In conclusion, this study provides compelling evidence that task fingerprinting offers a promising solution to overcome knowledge silos and enhance knowledge transfer in medical image analysis. Our newly proposed task distance measure bKLD demonstrates clear advantages across various knowledge transfer scenarios and exhibits large flexibility to be applied in a wide range of use cases. Future research is set to focus on expanding the framework's capabilities and exploring the broader impact on medical AI research.

\section{Methods}\label{sec:methods}

Our knowledge transfer framework is based on the following concept (Fig. \ref{fig:concept}): Any contribution by a participant of the network is associated with a task fingerprint, i.e., a shareable task representation that respects individual privacy regulation (e.g., that the data itself must not be shared). Meta information about suitable training design decisions, such as model architecture \rev{(e.g., a \textit{timm} \cite{rw2019timm} string)} or augmentation policy \rev{(e.g., an \textit{albumentations} \cite{Buslaev2020albumentations} serialization)}, is stored along\rev{side \sout{with}} the fingerprint. Optionally, the data may be tagged as shareable. To learn from the current network of tasks, a user generates the fingerprint for their task and queries the knowledge cloud. Based on the most relevant tasks in the pool according to fingerprint matching, relevant training strategies and data can be retrieved. The retrieved meta information and data are used to compile a training pipeline with different components of transferred knowledge. This section details our core methodological contributions as well as the evaluation concept we developed to assess the value of our approach. 

\subsection{Task fingerprinting}\label{ssec:meth:fingerprinting}

The major challenge of task fingerprinting is the efficient extraction of relevant task information into a fingerprint without compromising private information. Formally, we define a \textbf{task} $\mathfrak{T}$ as $\mathfrak{T} = \{(x_i, y_i)\}_{i=1}^{N}$ comprising $N$ samples, each consisting of an image $x_i$ and a label $y_i$, following Achille et al. \cite{achille2019information}. 
The process of \textbf{task fingerprinting} refers to computing a mapping $f$ of a task $\mathfrak{T}$ onto a set of real valued vector from within $\mathbb{R}^n$. Generated fingerprints of tasks must be comparable to measure similarity (and thus the corresponding potential of knowledge transfer) between tasks. This is done with the help of a \textbf{distance measure} $d$ (not necessarily symmetric or positive) that maps two fingerprints onto a real value ("the lower, the higher the similarity").

Our proposed fingerprinting method is visualized in Fig. \ref{fig:bkld}. To generate the fingerprint of a task $\mathfrak{T}$ we feed a fixed amount of samples ($n= 10,000$) through an ImageNet \cite{Deng2009ImageNetAL} pretrained ResNet34 \cite{He2016DeepRL} extracting $n$ deep feature vectors $\{\mathbf{v}_i \in \mathbb{R}^{m}\}_{i=1}^{n}$ from the activations ($m= 512$) prior to the classification head. Samples are drawn randomly with replacement and slight data augmentation (RandomCrop and HorizontalFlipping) is applied. Following the strategy of Ramtoula et al. \cite{Ramtoula2023VisualDR}, we distribute the activations of each feature dimension $\{1, .., m\}$ into $b$ uniform bins - creating $m$ activation histograms. After normalizing along the feature dimension the resulting feature distributions $\{\mathbf{p}_i \in [0,1]^{b}\}_{i=1}^{m}$ can be shared without revealing the features of individual samples.

To compare these fingerprints we use the Kullback-Leibler divergence \cite{Kullback1951OnIA}, which has been used for this purpose before \cite{Sugiyama2008DirectIE,Bhattacharjee2020P2LPT}, jointly with an optional weighting of features. The weights are hyperparameters that allow to shift focus on features that are particularly relevant for either the source or target task. Assume some weights $\mathbf{w} \in \mathbb{R}^m$ and tasks $\mathfrak{T}, \mathfrak{S}$ with $f(\mathfrak{T}) = \{\mathbf{p}_i \in [0,1]^{b}\}_{i=1}^{m}$ as well as $f(\mathfrak{S}) = \{\mathbf{q}_i \in [0,1]^{b}\}_{i=1}^{m}$. Then we define 

\begin{align*}
    d(f(\mathfrak{S}), f(\mathfrak{T})) &= \sum_{i=1}^m w^{(i)} \cdot \textbf{KLD} (p_i, \operatorname{soft}(q_i)) \\
    &= \sum_{i=1}^m w^{(i)} \cdot \sum_{j=1}^b p_i^{(j)} \cdot \log \frac{p_i^{(j)}}{\hat{q}_i^{(j)}},
\end{align*}

where $\operatorname{soft}$ is the \textbf{softmax} function such that $\hat{q}_i^{(j)} = \exp q_i^{(j)} / \sum_j \exp q_i^{(j)}$. The softmax is used to circumvent potentially empty bins. The key novelty of our approach is the combination of a sufficiently granular task embedding (compared to P2L \cite{Bhattacharjee2020P2LPT}) with a flexible weighting to guide feature attention (compared to VDNA \cite{Ramtoula2023VisualDR}).

\subsection{\rev{Hyperparameter optimization}}\label{ssec:meth:hyperparameters}

\rev{During the development phase, we began by exploring variations of previously proposed methods for task similarity \cite{Ramtoula2023VisualDR,achille2019task2vec,Ding2021AnalyzingDN,Bhattacharjee2020P2LPT,feydy2019interpolating,sun2016deep,Godau2021TaskFing}. We combined various embedding mappings $f$, including the binning scheme, with multiple distance measures $d$. We examined several smoothing functions (e.g., normalization, softmax, and symmetric uniform smoothing) for both the source and target embeddings, as well as weighting schemes. The list of all configurations can be found in our code repository.} Based on our exploratory experiments on the development tasks (see Fig. \ref{fig:data}), we identified \rev{\sout{three} two} favorable settings of hyperparameters: \textbf{bKLD(small\rev{\sout{,target}})} a "small" fingerprint with $b=100$ bins and weighting by the softened feature average of the target task $w = \operatorname{soft}(\sum_i^n \mathbf{v}_i / n)$\rev{\sout{, \textbf{bKLD(large,source)} with $b=1,000$ bins and weighting by normalized feature average of the source task}} as well as \textbf{bKLD(large\rev{\sout{,unweighted)}}} with $b=1,000$ and uniform weighting. To compare bKLD with the manual task matching approach, we extracted keywords from the semantic description of tasks, e.g., imaging modality, anatomical regions, and entities of interest, and computed the intersection over union between any pair of tasks. Ties were resolved by preferring larger source tasks.

\subsection{Knowledge transfer}\label{ssec:meth:knowledge}

A \textbf{task selector} is a function $s$, that given a (target) task $\mathfrak{T}$, and a set of potential source tasks $\mathbf{P}$ (not including $\mathfrak{T}$), returns a subset of $\mathbf{P}$ that are called candidates. For simplicity, we also allow $s$ to return a single candidate (instead of a singleton). Given a fingerprinting mapping $f$ and a distance measure $d$, the map $s_{f,d}(\mathfrak{T}) := \argmin_{\mathfrak{S} \in \mathbf{P}} d(f(\mathfrak{S}),f(\mathfrak{T}))$ becomes a task selector. 

Assume $t$ to be a \textbf{trainer} for a model on target task $\mathfrak{T}$ that defines all elements of the training pipeline (e.g., data augmentation, neural architecture, data sampling, ...), while potentially involving information from an additional source task $\mathfrak{S}$. 
The kind of knowledge transfer between source and target (e.g., pretraining, co-training) is not restricted in our framework. With $t_{\mathfrak{S}}$ we note the fully characterized training procedure of the model using $\mathfrak{S}$ as source task and $t$ as a trainer. For a performance metric $m$ that evaluates models on their target task based on a held-out test set, $m(t_{\mathfrak{S}}(\mathfrak{T}))$ describes the performance for task $\mathfrak{T}$ after the training of $t_{\mathfrak{S}}(\mathfrak{T})$ has taken place. As a baseline comparison $m(t_{\empty}(\mathfrak{T}))$ measures the performance for task $\mathfrak{T}$ after training a model with information solely given by $\mathfrak{T}$. Given $m$ is oriented positively ("the larger the better"), then we call the knowledge transfer "negative" if $m(t_{\mathfrak{S}}(\mathfrak{T})) < m(t_{\empty}(\mathfrak{T}))$ \cite{Wang2018CharacterizingAA}.

\subsection{Validation metrics}\label{ssec:meth:evaluation}

Following recent recommendations \cite{MaierHein2022MetricsRR}, we measured model performance with two complementary and broadly applicable \textbf{base metrics} $m$, namely Balanced Accuracy (BA) and (macro averaged) Area Under the Receiver Operator Characteristic (AUROC). In order to evaluate the quality of a task selector $s$ for a given target task $\mathfrak{T}$ upon a pool of potential source tasks $\mathbf{P}$, a variety of \textbf{meta metrics} exists. Assume $s$ chose source task $\mathfrak{S}$, and $m$, $t$ are given, let $\mathbf{O} :=\{m(t_{\mathfrak{S_i}}(\mathfrak{T})) | \mathfrak{S_i} \in \mathbf{P}\}$ be the set of all possible transfer outcomes, then:

\textbf{Improvement} measures the difference in applying the knowledge transfer via $s$ compared to the baseline: $m(t_{\mathfrak{S}}(\mathfrak{T})) - m(t_{\empty}(\mathfrak{T}))$. \textbf{Percentile} measures the relative position of $m(t_{\mathfrak{S}}(\mathfrak{T}))$ within all potential transfer outcomes: $|\{o \in \mathbf{O}: o \leq m(t_{\mathfrak{S}}(\mathfrak{T}))\}| / |\mathbf{O}|$. \textbf{Regret} \cite{Renggli2022WhichMT} measures how much of the remaining performance gap (from the performance determined by $s$ to optimal performance) could be overcome by making the optimal choice within $\mathbf{P}$: $[\max \mathbf{O} - m(t_{\mathfrak{S}}(\mathfrak{T}))] / [1 - m(t_{\mathfrak{S}}(\mathfrak{T}))]$. \textbf{Gain} \cite{Zamir2018TaskonomyDT} is a weaker form of improvement and returns 0 for a negative transfer, otherwise 1. 

Note that it is a realistic scenario to query for multiple shots of $s$ and retrieve the top $k$ candidates $\mathfrak{S}_1, ..., \mathfrak{S}_k$. For evaluation aggregation, one may either average over the individual results of selections and assess the expected quality of any suggestion (as we did in Fig. \ref{fig:comparison}) or proceed only with the best suggestion to assess the expected outcome (as we did in Fig. \ref{fig:benefits} and \ref{fig:robustness}). In contrast, \textbf{weightedtau} \cite{Agostinelli2022HowStable} considers the full ranking of candidates in $\mathbf{P}$. It is a weighted version of the Kendall rank correlation coefficient. We use a hyperbolic drop-off in importance as it is implemented in scikit-learn \cite{scikit-learn}. This allows us to compare the ranking of candidates by $s$ with the ranking of all potential sources via the actual outcomes $\mathbf{O}$.

\subsection{Data}\label{ssec:meth:data}

To guarantee broad applicability of our method, we gathered a very heterogeneous task pool from publicly available datasets. Figure \ref{fig:data} gives an overview of all used tasks, which amount to 71 in total. Inclusion criteria for datasets were (1) public availability, (2) use of 2D images, and (3) provided classification labels. A focus was laid on datasets from the medical domain (62 tasks), as well as a great variety in imaging domains (laparoscopy, dermatoscopy, gastroscopy, colonoscopy, ophthalmic microscopy, confocal laser endomicroscopy, MRI, X-ray, fundus photography, CT, capsule endoscopy, ultrasound). Supplementary Table \ref{tab:datasets} provides more details on the data sources \cite{Pogorelov2017,Borgli2020,Leibetseder2018LapGyn,Twinanda2017EndoNetAD,KhoslaYaoJayadevaprakashFeiFei_FGVC2011,Deng2009ImageNetAL,Netzer2011ReadingDI,FeiFei2004LearningGV,Griffin2007Caltech256OC,Krizhevsky2009LearningML,Faria2019LightFI,Kawahara2019SevenPointCA,LeCun1998GradientbasedLA,Cohen2017EMNISTEM,sara_moccia_2018_1162784,sara_moccia_2017_1003200,irvin2019chexpert,KERMANY20181122,Jaeger2014TwoPC,HematejaAluru2021,Pranav2017,beansdata,Negin2020,barrets,ALDHABYANI2020104863,jr2ngb,Bohaju2020brain,cheng2015enhanced,Yang2020COVIDCTDatasetAC,Giotis2015mednode,ferreira2015ph2,isic20dataset,Rotemberg2021APD,LIU2022100512,aptos2019,Smedsrud2021}. Each task was split into train and test sets ensuring proportions of 80:20 and equal class distributions between splits. All images were preprocessed to match a resolution of $256 \times 256$ pixels. Greyscale images were further transferred to RGB. For tasks that exceeded 1,000 samples, we subsampled a shrunk version of 800 train samples (following \cite{Renggli2022WhichMT}) that served as a target task variant. This served to reduce compute resources, increase comparability across target tasks, and allow knowledge transfer to have a significant influence as, e.g., pretraining is the most advantageous in scenarios where the target task $\mathfrak{T}$ is small \cite{Raghu2019transfusion}. This implies that all task distances are estimated from a "complete" source task to a "shrunk" target task. We partitioned the set of all tasks into a development set $\mathbf{P}_{\text{dev}}$ and a validation set $\mathbf{P}_{\text{val}}$. We ensured that all tasks related to the tasks we used in our previous study \cite{Godau2021TaskFing} were selected as development tasks. Tasks from the validation split were masked for the process of hyperparameter selection for bKLD. Note that there is a severe (imaging) domain shift from development to validation tasks, challenging the generalizability of our task selector. Furthermore, note the high variability in task size (min:170, q1:1572, q3:40'673, max:122'138), number of classes (min:2, q1:2, q3:5, max:257), imbalance ratio (min:1, q1:1.34, q3:14.27, max:171.33), and resolutions. To the best of our knowledge, no imaging task similarity measure has been evaluated this extensively before.

\subsection{Validation of knowledge transfer quality}\label{ssec:meth:experiments}

The purpose of our experiments was to demonstrate the benefit of task fingerprinting for a variety of knowledge transfer scenarios. We therefore designed four realistic scenarios, covering both cases with only source task metadata available ("Model Architecture" and "Augmentation Policy") as well as cases with source task data available ("Pretraining Data" and "Co-Training Data"). For the former, we created separate knowledge clouds and conducted initial experiments as described below. As a baseline, we also trained models on all (shrunk) target tasks in isolation. During our search for optimal bKLD hyperparameters, we fixed $\mathbf{P}_{\text{dev}}$ both as a pool of source and target tasks. For the final evaluation, we set $\mathbf{P}_{\text{val}}$ as a pool of target tasks and allowed both development and validation tasks as a pool of available source tasks. In any case of image overlap between two tasks, we excluded those tasks as respective potential source tasks (the DeepDRiD \cite{LIU2022100512}, Cholec80 \cite{Twinanda2017EndoNetAD}, and CheXpert \cite{irvin2019chexpert} tasks).

\rev{To facilitate comparison, we included a selection of previously proposed task selectors based on task distances as baselines. These were chosen  to balance “classic” methods, such as FID \cite{Heusel2017FID,Ding2021AnalyzingDN}), with more recent methods, such as VDNA \cite{Ramtoula2023VisualDR}). We also compared bKLD with approaches that use more dense task representations, such as Sinkhorn Divergence \cite{feydy2019interpolating}, as well as approaches that use much simpler representations, such as P2L \cite{Bhattacharjee2020P2LPT}). FED \cite{Godau2021TaskFing,achille2019task2vec} complements these baselines because it is not based on feature extraction. The “manual” baseline was intended to capture a common and tangible approach to the problem.}

\subsubsection{Single task baseline}
The single task baseline comprises a “transfer-free” training recipe and subsequently simulates a model performance that is indicative of an achievable result without significant resource investment in experimentation. Coming up with such a fair and meaningful training scheme for each task presents a major challenge given the large heterogeneity between them. Too much individual optimization per task is costly and introduces another potential factor of bias; on the other hand, applying a uniform training scheme on all tasks might lead to unrealistic performances. We aimed for a balance with the following strategy: For faster convergence and more stable training, we use ImageNet \cite{Deng2009ImageNetAL} pretrained models throughout \cite{Raghu2019transfusion}. Starting from the recommendations given by Wightman et al. \cite{Wightman2021ResNetSB}, we made slight adaptations to the training pipeline to ensure model convergence, avoid overfitting, and generally improve performance on $\mathbf{P}_{\text{dev}}$. On an individual task level, we used the automated tuning of the learning rate \cite{Smith2017CyclicalLR} that is implemented in Pytorch Lightning \cite{Falcon_PyTorch_Lightning_2019}. \rev{Medical data often suffers from class imbalance, which can affect model performance. To address this important issue, we worked with a variety of imbalanced tasks, measuring imbalance as the ratio of the size of the largest class to the size of the smallest class, with ratios up to 171. Class-balanced sampling was used to show the models the less frequent classes with a higher probability.} \rev{The pipeline was optimized collectively across tasks during development. Afterwards, it was validated on the validation tasks. It is worth noting that this approach occasionally results in non-convergence during training and frequently performs less effectively than a training procedure that incorporates more task-specific pipeline tuning. However, we argue that the primary purpose of task fingerprinting is not to immediately produce a state-of-the-art model, but rather to refine the  model development process by utilizing advantageous initialization and assisted iteration. Second, in order to validate the methodology, it is sufficient to demonstrate positive trends in a suboptimal yet sensible regime.} As soon as we identified a solid configuration, we kept it for all ensuing experiments. The unmodified pipeline served as a baseline to compute gain and improvement. To compute the other meta metrics, it was necessary to train all potential knowledge transfer-induced pipeline adaptations (on all tasks) - this allowed us to rank all source tasks (percentile and weightedtau) and deduce the theoretically optimal performance (regret).

\subsubsection{Scenario 1: Model Architecture}
The first knowledge transfer scenario concerns the transfer of the neural architecture (including its pretrained weights), which has previously been suggested in literature \cite{Dias2021ImageDataset2VecAI, Godau2021TaskFing}. Given a candidate source task $\mathfrak{S}$ and the attached meta information $m(\mathfrak{S})$ for multiple neural architectures, $t_{\mathfrak{S}}(\mathfrak{T})$ refers to using the architecture (and initialization) with the best performance $m(\mathfrak{S})$ on the source task $\mathfrak{S}$ for training target task $\mathfrak{T}$. To generate the necessary meta information for each potential source task, we fixed a set of 20 candidate architectures, based on the following criteria: (1) availability within the PyTorch Image Models \cite{rw2019timm}, (2) a reported ImageNet \cite{Deng2009ImageNetAL} accuracy above 75\% according to PyTorch Image Models \cite{rw2019timm} while using at most 30 million parameters, which is to allow for (3) the applicability of the architecture with our compiled training pipeline within the restrictions of our hardware (max. GPU VRAM of 24GB). Finally, we reduced the list to architectures (4) published from 2019 onwards to capture the current state-of-the-art. For architectures available in different sizes, we chose the single largest variant that met our hardware requirements. The list of resulting 20 neural architectures \cite{Xie2020SelfTrainingWN,Zhang2020ResNeStSN,Wang2020ECANetEC,Yalniz2019BillionscaleSL,Xie2020AdversarialEI,Tan2019EfficientNetRM,Bochkovskiy2020YOLOv4OS,Wang2020CSPNetAN,Lee2019AnEA,Tan2019MixConvMD,Radosavovic2020DesigningND,Gao2021Res2NetAN,Han2021RethinkingCD,Yang2019CondConvCP,Li2019SelectiveKN,Sun2019HighResolutionRF,Howard2019SearchingFM,He2016DeepRL} is presented in Supplementary Table \ref{tab:timm_architectures}. Note that all neural architectures have at least one source task, they perform best on (in contrast to \cite{Dias2021ImageDataset2VecAI}).

\subsubsection{Scenario 2: Pretraining Data}

The second knowledge transfer scenario involves the commonly applied concept of fine-tuning. Given a source task $\mathfrak{S}$, the training of $t_{\mathfrak{S}}(\mathfrak{T})$ consists of a pretraining phase, during which a model is trained on the task $\mathfrak{S}$, followed by fine-tuning (with a newly attached final classification layer) on the target task $\mathfrak{T}$. This setup has been used in previous work \cite{Bhattacharjee2020P2LPT,Zamir2018TaskonomyDT,Agostinelli2022HowStable}.

\subsubsection{Scenario 3: Augmentation Policy}

The third knowledge transfer scenario centers around automatically learned data augmentation \cite{Cubuk2018AutoAugmentLA,Hataya2020FasterAL}. Optimal data augmentation is an important ingredient for training state-of-the-art models \cite{Wightman2021ResNetSB}, but remains highly task dependent \cite{Yang2022ImageDA,Shorten2019ASO}. The automation on this part of the training pipeline offers potential benefits especially in the low data regime, when augmentation is necessary to avoid model overfitting \cite{Shorten2019ASO}. Nevertheless the process of automatically generating augmentation policies is highly resource intense \cite{Cubuk2018AutoAugmentLA} even with proposed speed improvements \cite{Hataya2020FasterAL}. Therefore, the transfer and reuse of learned policies have been suggested \cite{Cubuk2018AutoAugmentLA}. In our experimental setup, this is modeled as follows: For each potential source task $\mathfrak{S}$, we compute a task-specific augmentation policy according to Albumentations' \cite{Buslaev2020albumentations} implementation of FasterAutoAugment \cite{Hataya2020FasterAL}, and attach it as meta information. $t_{\mathfrak{S}}(\mathfrak{T})$ refers to the transfer of the corresponding augmentation policy of $\mathfrak{S}$ to the training of target task $\mathfrak{T}$.

\subsubsection{Scenario 4: Co-Training Data}

The fourth knowledge transfer scenario focuses on the immediate interplay of source and target tasks during neural network training. The research around multi-task learning  \cite{Crawshaw2020MultiTaskLW} has revealed interesting insights regarding this interplay: Optimal source tasks for pretraining might differ from optimal learning partners in multi-task learning \cite{Standley2020WhichTS}. We thus explicitly include this setup as a complementary scenario, which has so far received only moderate attention in the literature on task similarity \cite{Zhou2021mtl,Fifty2021EfficientlyIT}. Given source and target tasks $\mathfrak{S}$ and $\mathfrak{T}$, the procedure for $t_{\mathfrak{S}}(\mathfrak{T})$ involves attaching two separate classifier heads to the shared backbone of the neural network. Training samples are drawn with equal likelihood from each of the two tasks $\mathfrak{S}$ and $\mathfrak{T}$, while only the corresponding samples are used to compute the respective losses for each of the heads. Both losses are weighted equally before backpropagation.

\subsubsection{\rev{Generalization Scenario}}

\rev{
To investigate the generalization of our methodology to different modalities and tasks prospectively, we applied the concept developed on 2D classification tasks to 3D segmentation tasks. To this end, we leveraged one of the most widely used datasets, namely the Medical Segmentation Decathlon challenge data~\cite{simpson2019large}, and official results~\cite{antonelli2022medical}. For feature extraction we used a Masked Autoencoder (MAE)~\cite{he2022masked} pretrained by Eckstein et al.~\cite{eckstein2025missing} on two large 3D datasets~\cite{hamamci2024developing,casey2018adolescent} based on the implementation by Wald et al.~\cite{wald2024openmind} and a ResEncL backbone~\cite{isensee2024nnu}. Again, we extracted 10,000 feature maps per dataset from the bottleneck layer of the network, without masking the input patch. The feature maps originally had a size of $320 \times 5 \times 5 \times 5$, however for computational efficiency, we reduced all but the first dimension via the arithmetic mean. The patches were sampled randomly across patients, available imaging modalities, and spatial locations. The challenge provides the validation values for 19 participating teams, on 17 tasks for two performance measures: Dice Similarity Coefficient (DSC) and Normalized Surface Dice (NSD). When analyzing the results, however, we observed that the distribution of "best models per task" was skewed (a single participant won 29 out of the 34 rankings). Hence, applying our methodology for scenario 1, which only considers the best performance on the source task, would lead to a trivial setup, where almost all source tasks would be optimal for most target tasks. To address this issue, we computed the transferability between tasks as the Pearson correlation coefficient of the participants’ results on two target tasks. Note that we excluded two participants who occasionally performed with a value of zero, indicating  algorithmic failure.}

\backmatter

\bmhead{Acknowledgements}

The authors would like to thank Maike Rees for her help on designing Fig. \ref{fig:concept} and Fig. \ref{fig:data}. We would also like to thank all members of the "Intelligent Systems in Surgical Endoscopy" team for fruitful discussions. We are grateful for the linguistic corrections by Minu Dietlinde Tizabi and Marcel Knopp. This project has been funded by the German Federal Ministry of Health under the reference number 2520DAT0P1 as part of the
pAItient (Protected Artificial Intelligence Innovation Environment for Patient Oriented Digital Health Solutions for developing, testing and evidence based evaluation of clinical value) project. The funder played no role in study design, data collection, analysis and interpretation of data, or the writing of this manuscript. The present contribution is supported by the Helmholtz Association under the joint research school
“HIDSS4Health – Helmholtz Information and Data Science School for Health". \rev{This project has received funding from the European Research Council (ERC) under the European Union’s Horizon 2020 research and innovation programme (project NEURAL SPICING grant agreement No. 101002198).}

\section*{Declarations}
\subsection*{Conflict of interest}
All authors declare no financial or non-financial competing interests. 
\subsection*{Data availability}
All datasets used for this study are publicly available. Details can be found in Supplementary Table \ref{tab:datasets}. Intermediate results such as each performance measure of knowledge transfer experiments and all pairwise task distances, are provided in our code repository.
\subsection*{Code availability}
The source code can be found at \url{https://github.com/IMSY-DKFZ/task-fingerprinting} under the MIT license.
\subsection*{Author contribution}
P.G. and L.M.H. conceptualized the work. P.G., A.S.,\rev{C.U.,} T.A.\rev{, K.M.H.,} and L.M.H. designed the study. P.G.\rev{, C.U.,} and A.S. performed the coding and data preparation. P.G. trained all deep learning models and conducted the statistical analysis. T.A., \rev{K.M.H.,} and L.M.H. provided advisory support for the project. P.G. and L.M.H. prepared the initial draft of the manuscript, with all authors participating in the refinement.

\begin{appendices}

\end{appendices}


\bibliography{sn-references}

\newpage
\section{Supplementary information}\label{sec:app:extended}

We provide details and references for the used neural architectures (see Supplementary Table \ref{tab:timm_architectures}) and datasets (see Supplementary Table \ref{tab:datasets}). Supplementary Table \ref{tab:scenario_similarity} displays the weightedtau of measured transferability outcomes between knowledge transfer scenarios. Supplementary Table \ref{tab:win_rates} presents the win rates across compared task selectors and investigated knowledge transfer scenarios. In addition we provide the complementary approach of "rank then mean" to Fig. \ref{fig:comparison} of the main paper in Supplementary Fig. \ref{fig:app_comparison}.

\begin{table*}[p]
\renewcommand{\arraystretch}{1.3}
\caption{Overview on Neural Architectures used for model architecture transfer experiments. ImageNet Accuracy has been provided by the timm library \cite{rw2019timm}. Train parameters only comprise the shared backbone.}
\label{tab:timm_architectures}
\centering
\begin{tabular}{|c|c|c|c|}
\hline 
\textbf{Architecture} & \textbf{timm reference} & \textbf{ImageNet Acc. (\%)} & \textbf{train params (mio.)} \\
\hline
EfficientNet B2 noisy student \cite{Xie2020SelfTrainingWN} & tf\_efficientnet\_b2\_ns     & 82.38  & 7.7  \\
ResNet50 SWSL \cite{Yalniz2019BillionscaleSL} & swsl\_resnet50               & 81.166 & 23.5 \\
ResNeSt50 \cite{Zhang2020ResNeStSN} & resnest50d                   & 80.974 & 25.4 \\
ECA ResNet50 \cite{Wang2020ECANetEC} & ecaresnet50d                 & 80.592 & 23.5 \\
ResNeXt50 SSL \cite{Yalniz2019BillionscaleSL} & ssl\_resnext50\_32x4d        & 80.318 & 23.0 \\
EfficientNet B2 AdvProp \cite{Xie2020AdversarialEI} & tf\_efficientnet\_b2\_ap     & 80.3   & 7.7  \\
EfficientNet B2 \cite{Tan2019EfficientNetRM} & tf\_efficientnet\_b2         & 80.086 & 7.7  \\
CSP DarkNet53 \cite{Bochkovskiy2020YOLOv4OS} & cspdarknet53                 & 80.058 & 26.6 \\
CSPResNeXt50 \cite{Wang2020CSPNetAN} & cspresnext50                 & 80.04  & 18.5 \\
CSPResNet50 \cite{Wang2020CSPNetAN} & cspresnet50                  & 79.574 & 20.6 \\
VoVNet \cite{Lee2019AnEA} & ese\_vovnet39b               & 79.32  & 23.5 \\
MixNet-L \cite{Tan2019MixConvMD} & mixnet\_l                    & 78.976 & 5.8  \\
RegNetY \cite{Radosavovic2020DesigningND} & regnety\_032                 & 78.886 & 17.9 \\
RegNetX \cite{Radosavovic2020DesigningND} & regnetx\_032                 & 78.172 & 14.3 \\
Res2Net50 \cite{Gao2021Res2NetAN} & res2net50\_26w\_4s           & 77.964 & 23.7 \\
RexNet100 \cite{Han2021RethinkingCD} & rexnet\_100                  & 77.858 & 3.5  \\
EfficientNet B0 CondCov \cite{Yang2019CondConvCP} & tf\_efficientnet\_cc\_b0\_4e & 77.306 & 12.0 \\
SK ResNet34 \cite{Li2019SelectiveKN} & skresnet34                   & 76.912 & 21.8 \\
HRNetV2-W18 \cite{Sun2019HighResolutionRF} & hrnet\_w18                   & 76.758 & 19.3 \\
MobileNetV3-Large \cite{Howard2019SearchingFM} & mobilenetv3\_large\_100      & 75.766 & 4.2 \\
\hline
ResNet34 \cite{He2016DeepRL} & resnet34 & 75.11 & 21.3 \\
\hline 
\end{tabular}
\end{table*}

\begin{table*}[p]
\renewcommand{\arraystretch}{1.3}
\centering
\caption{Overview on datasets and tasks used within out task pool. More properties can be found in Fig. \ref{fig:data}.}
\label{tab:datasets}
\begin{tabular}{|c|c|c|}
\hline 
\textbf{Dataset} & \textbf{Tasks} & \textbf{Reference(s)}\\
\hline 
Nerthus & T01 &  \cite{Pogorelov2017} \\
HyperKvasir & T02-T05 & \cite{Borgli2020}  \\
LapGyn4 & T06-T09 &  \cite{Leibetseder2018LapGyn,Twinanda2017EndoNetAD} \\
Cholec80 & T10-T16 &  \cite{Twinanda2017EndoNetAD} \\
Stanford Dogs & T17 & \cite{KhoslaYaoJayadevaprakashFeiFei_FGVC2011,Deng2009ImageNetAL} \\
SVHN & T18 &  \cite{Netzer2011ReadingDI} \\
Caltech101 & T19 &  \cite{FeiFei2004LearningGV} \\
Caltech256 & T20 &  \cite{Griffin2007Caltech256OC} \\
CIFAR & T21, T22 &  \cite{Krizhevsky2009LearningML} \\
SKLIN2 & T23 &  \cite{Faria2019LightFI} \\
Derm7pt & T24 &  \cite{Kawahara2019SevenPointCA} \\
MNIST & T25 & \cite{LeCun1998GradientbasedLA} \\
EMNIST & T26 &  \cite{Cohen2017EMNISTEM} \\
NBI-InfFrames & T27 & \cite{sara_moccia_2018_1162784} \\
Laryngeal Cancerous Tissues & T28 &  \cite{sara_moccia_2017_1003200} \\
\hline
CheXpert & T29-T41 & \cite{irvin2019chexpert} \\
Zhang Chest X-Ray Images  & T42 & \cite{KERMANY20181122} \\
Shenzen Hospital CXR Set  & T43 & \cite{Jaeger2014TwoPC} \\
Kaggle COVID X-Ray Dataset & T44 & \cite{HematejaAluru2021} \\
MURA & T45-T51 &  \cite{Pranav2017} \\
ibean & T52 &  \cite{beansdata} \\
CatRelComp & T53 & \cite{Negin2020} \\
AIDA-E Barrett’s Esophagus Diagnosis & T54 &  \cite{barrets} \\
Dataset of Breast Ultrasound Images & T55 & \cite{ALDHABYANI2020104863} \\
Kaggle Cataract Dataset & T56 &  \cite{jr2ngb} \\
Kaggle Brain Tumor Classification & T57 &  \cite{Bohaju2020brain} \\
Brain Tumor Type Classification & T58 &  \cite{cheng2015enhanced} \\
COVID-CT-Dataset & T59 &  \cite{Yang2020COVIDCTDatasetAC} \\
MED-NODE  & T60 &  \cite{Giotis2015mednode} \\
PH2 & T61 &  \cite{ferreira2015ph2} \\
ISIC20 & T62 &  \cite{isic20dataset,Rotemberg2021APD} \\
DeepDRiD & T63-T67 & \cite{LIU2022100512} \\
APTOS19 & T68 & \cite{aptos2019} \\
Kvasir-Capsule & T69-T71 & \cite{Smedsrud2021} \\
\hline
\end{tabular}
\end{table*}

\begin{table*}[p]
\renewcommand{\arraystretch}{1.3}
\caption{Mean and standard deviation of weightedtau on pairwise transfer scenario outcomes over 43 tasks, 3 repetitions and 3 performance metrics for the four knowledge transfer scenarios we investigate: Model Architecture (M. A.), Pretraining Data (P. D.), Augmentation Policy (A. P.) and Co-Training Data (C. D.).}
\label{tab:scenario_similarity}
\centering
\begin{tabular}{|c|c|c|c|c|}
\hline 
  & \textbf{Model Architecture} & \textbf{Pretraining Data} & \textbf{Augmentation Policy} & \textbf{Co-Training Data} \\
\hline
\textbf{M. A.} & 1.000 ± 0.000 & 0.084 ± 0.147 & 0.078 ± 0.141  & 0.052 ± 0.137  \\
\textbf{P. D.} & 0.084 ± 0.147 & 1.000 ± 0.000 & 0.037 ± 0.164  & 0.038 ± 0.136  \\
\textbf{A. P.} & 0.078 ± 0.141 & 0.037 ± 0.164 & 1.000 ± 0.000  & -0.002 ± 0.140 \\
\textbf{C. D.} & 0.052 ± 0.137 & 0.038 ± 0.136 & -0.002 ± 0.140 & 1.000 ± 0.000 \\
\hline 
\end{tabular}
\end{table*}

\begin{table*}[p]
\renewcommand{\arraystretch}{1.3}
\caption{Win rates \cite{Agostinelli2022HowStable} for task selectors described in Fig. \ref{fig:comparison} in percent. Shows the fraction of 1032 individual cases (43 tasks, 3 repetitions, 4 meta metrics, 2 base metrics) that a specific task selector performs best. Columns may sum above 100 because of ties, to reduce such occurrences we averaged the meta metrics of top 3 suggestions. Next to the win rates for four knwoledge transfer scenarios we also provide the average across them.}
\label{tab:win_rates}
\centering
\begin{tabular}{|c|c|c|c|c|c|}
\hline 
  & \textbf{M. A.} & \textbf{P. D.} & \textbf{A. P.} & \textbf{C. D.} & \textbf{mean}\\
\hline
\textbf{FID}                      & 15.02 & 13.86 & 12.60 & 10.66 & 13.03 \\
\textbf{P2L}               & 15.21 & 14.92 & 13.66 & 10.17 & 13.49 \\
\textbf{FED}                      & 15.89 & 15.21 & 11.14 & 12.79 & 13.76 \\
\textbf{VDNA}     & 13.66 & 18.41 & 16.09 & 12.21 & 15.09 \\
\textbf{Manual}                 & 20.64 & 19.86 & 11.34 & 16.96 & 17.20 \\
\textbf{bKLD(small,target)}  & 14.53 & \textbf{25.58} & 23.84 & 17.25 & 20.30 \\
\textbf{bKLD(large,unweighted)}      & 14.53 & 25.39 & 22.19 & 20.64 & 20.69 \\
\textbf{bKLD(large,source)} & \textbf{34.98} & 19.86 & \textbf{27.71} & \textbf{32.66} & \textbf{28.80} \\
\hline 
\end{tabular}
\end{table*}

\begin{figure*}[p]
\centering
\includegraphics[width=\textwidth]{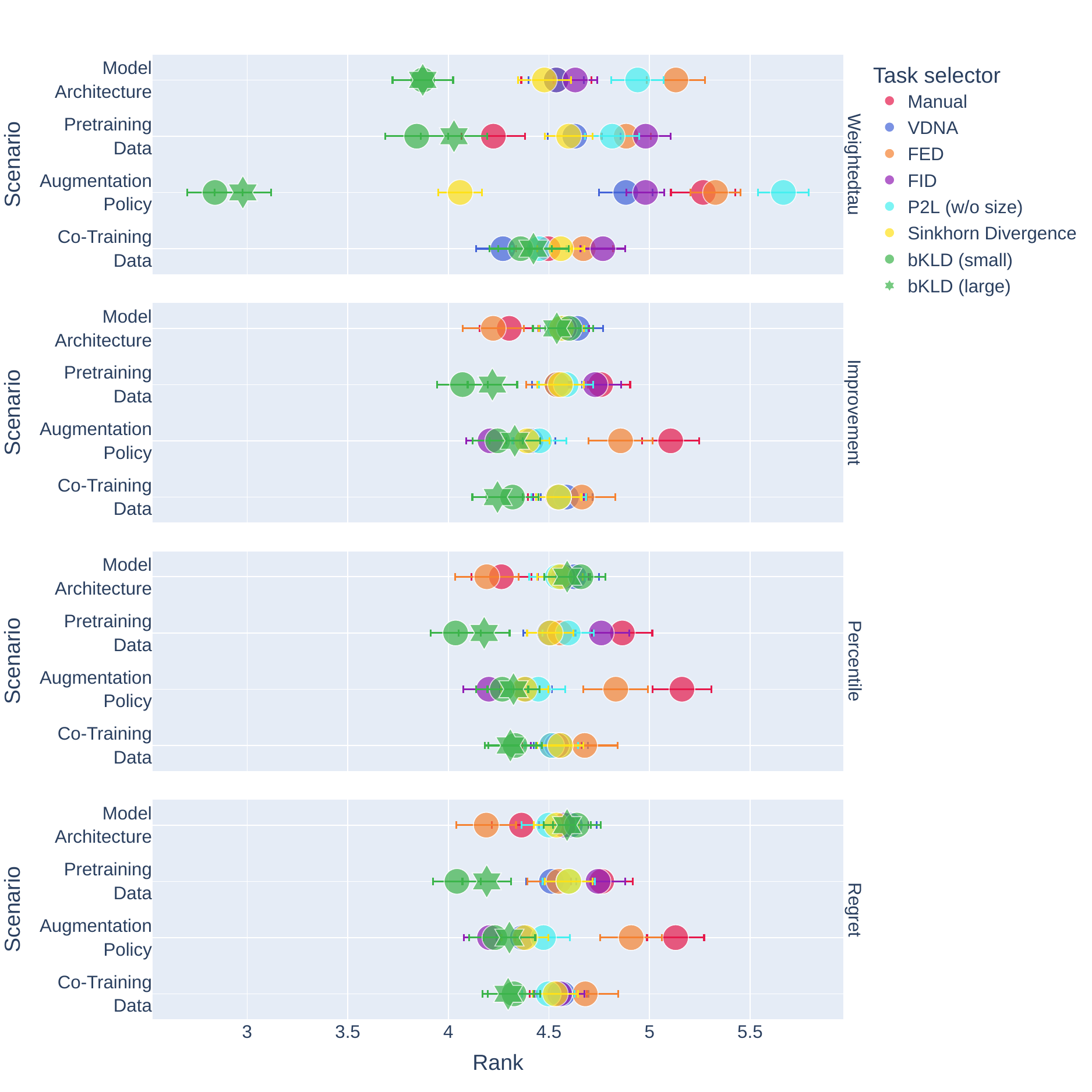}
\caption{\textbf{binned Kullback-Leibler Divergence (bKLD) outperforms previously proposed methods for knowledge transfer.} In contrast to the evaluation shown in Fig. \ref{fig:comparison}, this figure shows the evaluation according to “rank then mean” \cite{Wiesenfarth2021RR}, using 258 setups (2 base metrics, 43 validation tasks, 3 repetitions). The marker position refers to the mean over 1000 bootstraps, with the whiskers indicating standard deviation. For each setup, the improvement, percentile, and regret of the top three suggestions are averaged, while weightedtau is evaluated on the full ranking of suggested knowledge transfer sources.}\label{fig:app_comparison}
\end{figure*}

\end{document}